\documentclass[conference]{IEEEtran}
\IEEEoverridecommandlockouts
\usepackage{cite}
\usepackage{amsmath,amssymb,amsfonts}
\usepackage[perpage]{footmisc}
\usepackage{algorithmic}
\usepackage{subcaption,graphicx}
\usepackage{caption}
\usepackage{textcomp}
\usepackage{xcolor}
\usepackage{multirow}
\usepackage{enumitem}
\usepackage{xcolor}
\usepackage{hyperref}
\usepackage{footnote}
\usepackage{float}
\usepackage{soul}
\usepackage{dblfloatfix}    
\hypersetup{colorlinks=true,linkbordercolor=red,linkcolor=blue,pdfborderstyle={/S/U/W 1}}

\begin{document}
\title{Mcity Data Collection for Automated Vehicles Study}
\author{\IEEEauthorblockN
{Yiqun Dong, Yuanxin Zhong, Wenbo Yu, Minghan Zhu, Pingping Lu, Yeyang Fang, Jiajun Hong, Huei Peng}
\IEEEauthorblockA{
\textit
{Department of Mechanical Engineering, University of Michigan, Ann Arbor, MI, USA, 48109}\\
\textbf
{\{yiqund, zyxin, wenboyu, minghanz, pingpinl, yeyangf, jiajunh, hpeng\}@umich.edu}
}}
\maketitle

\begin{abstract}
The main goal of this paper is to introduce the data collection effort at Mcity targeting automated vehicle development. We captured a comprehensive set of data from a set of perception sensors (Lidars, Radars, Cameras) as well as vehicle steering/brake/throttle inputs and an RTK unit. Two in-cabin cameras record the human driver's behaviors for possible future use. The naturalistic driving on selected open roads is recorded at different time of day and weather conditions. We also perform designed choreography data collection inside the Mcity test facility focusing on vehicle to vehicle, and vehicle to vulnerable road user interactions--which is quite unique among existing open-source datasets. The vehicle platform, data content, tags/labels, and selected analysis results are shown in this paper\footnote{Demonstrative examples can be found at: \url{https://drive.google.com/drive/folders/1GxZlwhymCl7MUiD30QfYlI49yUSXLjZG}}.  
\end{abstract}


\section{Introduction}

Automated vehicles (AVs) have the potential to radically impact our society\cite{societal_impact} by improving safety, congestion and energy consumption. Reliable AV operations require reliable sensing and  perception of the surrounding environment, e.g., to understand the presence and future motions of road users and the governing traffic rules.  Robust perception is the basis of safe/proper trajectory planning and control. To achieve reliable perception, deep neural networks are frequently used, which require large sets of data. In recent years, many open datasets were created and shared, first from universities and more recently, from companies.  Not all datasets include all three of the common AV sensor types and the tags/labels vary considerably among those datasets.

All three types of commonly used AV sensors (cameras, lidars and radars) have strength and weakness.  In addition, even within the lidar family, the mechanical scanning Velodyne lidar we used has 32 beams and covers much wider horizontal and vertical field of view, while the Ibeo lidar only has 4 beams and a limited field of view.  Because both lidar sensors are widely used in automotive applications but commonly for different purposes (e.g., Level 4 vs. Level 2), compare/contrast their performance is of interest \cite{dingzhao_multiple_lidar}.     

In the past, vehicle controls were largely designed based on model-based algorithms through mathematically rigorous processes \cite{Peng_preview_control, dong_control_1,dong_control_2}. Recent advances, however, have indicated the potential of data-driven approaches \cite{dingzhao_driving_behavior, xianan_paper}, which requires a large amount of training (and validation) data. In the past, quite a few open datasets were published, which help to elevate the state of the art of the data-driven approaches tremendously.  Nevertheless, many of them seem to only capture naturalistic driving, i.e., not deliberately focusing on challenging scenarios.  Based on our previous work on accelerated evaluation, we believe the challenging driving behaviors should be emphasized more.  In other words, collecting data naturalistically is time-consuming and costly.  A deliberate, choreography-designed set of scenarios conducted inside a safe and closed test facility can provide a different and useful set of data that is complementary to naturalistic data.

The overall guiding principle of our data collection effort is \textit{completeness}, including to deploy a wide set of sensors, cover a wide array of weather and lighting conditions, diverse lane marking on diverse road topology, and situations involving challenge interactions from other road users (vehicles, bicycles, pedestrians). In addition to collect naturalistic data on open roads, we also capture designed scenarios inside the Mcity test facilities, with the focus on intersections.  

Table \ref{dataset_comparison_big_table} compares our dataset with several other open datasets. Note that several commercial entities published some data recently, but some have very restrictive terms of use (e.g., Waymo\cite{waymo_dataset}), which we choose not to include in our comparison.  We summarize our contributions below:

\begin{itemize}[leftmargin = 9pt]
    \item \textbf{Controlled diversity}: We repeatedly collect naturalistic driving data on fixed routes with deliberate variation in lighting, weather, traffic, and human driver characteristics.  
    \item \textbf{Designed choreography}: We designed representative urban driving scenarios with the host vehicle interacting with other vehicle/pedestrian/cyclist inside the Mcity test facility. The test case parameters were selected to cover both normal (courteous, law-abiding) and abnormal (aggressive, against traffic law) conditions.    
    \item \textbf{Completeness}: As shown in Table \ref{dataset_comparison_big_table}, we use a comprehensive set of sensors.  
\end{itemize}

\begin{table*}[t]
    \centering
    \caption{Comparison of existing open datasets}
    \label{dataset_comparison_big_table}
    \begin{tabular}{ l c l c c c c c c c c c c}
    \hline
    \hline 
Dataset &Year &Locations &  \begin{tabular}[c]{@{}l@{}} \shortstack{Size\\(hr/mi)}  \end{tabular}   & \begin{tabular}[c]{@{}l@{}} \shortstack{Labeled\\Frames}\end{tabular}  & \begin{tabular}[c]{@{}l@{}} \shortstack{360$^o$ FOV\\Lidar}  \end{tabular} &\begin{tabular}[c]{@{}l@{}} \shortstack{Limited FOV\\Lidar}  \end{tabular} &Radar & \begin{tabular}[c]{@{}l@{}} \shortstack{Lighting\\Diversity} \end{tabular} & \begin{tabular}[c]{@{}l@{}} \shortstack{Weather\\Diversity} \end{tabular} & \begin{tabular}[c]{@{}l@{}}  \shortstack{Driving\\Behavior} \end{tabular} & \begin{tabular}[c]{@{}l@{}} \shortstack{Designed\\Choreography} \end{tabular} \\
\hline
Mcity dataset                   &2019     &AA, Mcity        &50/3k     &17.5k    &$\bullet$      &$\bullet$   &$\bullet$    &$\bullet$    &$\bullet$     &$\bullet$  &$\bullet$  \\\hline
Lyft\cite{lyft_dataset}         &2019     &CA               &-/-       &55k      &$\bullet$      &$\circ$     &$\circ$      &$\bullet$    &$\bullet$     &$\circ$    &$\circ$    \\\hline 
nuScenes\cite{nuscenes}         &2019     &Boston, SG       &5.5/55    &40k      &$\bullet$      &$\circ$     &$\bullet$    &$\bullet$    &$\bullet$     &$\circ$    &$\circ$    \\\hline 
H3D\cite{h3d}                   &2019     &CA               &0.77/-    &27k      &$\bullet$      &$\circ$     &$\circ$      &$\circ$      &$\circ$       &$\circ$    &$\circ$    \\\hline
HDD\cite{honda_dataset}         &2018     &CA               &104/-     &0        &$\bullet$      &$\circ$     &$\circ$      &$\circ$      &$\circ$       &$\circ$    &$\circ$    \\\hline
AS\cite{appolo}                 &2018     &China            &100/-     &144k     &$\circ$        &$\circ$     &$\circ$      &$\bullet$    &$\circ$       &$\circ$    &$\circ$    \\\hline 
AS lidar\cite{aslidar}          &2018     &China            &2/-       &20k      &$\bullet$      &$\circ$     &$\circ$      &$\circ$      &$\circ$       &$\circ$    &$\circ$    \\\hline 
KAIST\cite{kaist}               &2018     &Seoul            &-/-       &8.9k     &$\bullet$      &$\circ$     &$\circ$      &$\bullet$    &$\circ$       &$\circ$    &$\circ$    \\\hline 
Vistas\cite{mapillary}          &2017     &6 continents     &-/-       &25k      &$\circ$        &$\circ$     &$\circ$      &$\bullet$    &$\bullet$     &$\circ$    &$\circ$    \\\hline
BDD100k\cite{bdd}               &2017     &NY, SF           &1k/-      &100k     &$\circ$        &$\circ$     &$\circ$      &$\bullet$    &$\bullet$     &$\bullet$  &$\circ$    \\\hline 
Cityscapes\cite{Cityscapes}     &2016     &50 cities        &-/-       &25k      &$\circ$        &$\circ$     &$\circ$      &$\circ$      &$\circ$       &$\circ$    &$\circ$    \\\hline
RobotCar\cite{oxford}           &2015     &Oxford           &210/620   &0        &$\circ$        &$\bullet$   &$\circ$      &$\bullet$    &$\bullet$     &$\circ$    &$\circ$    \\\hline
KITTI\cite{kitti}               &2012     &Karlsruhe        &1.5/-     &15k      &$\bullet$      &$\circ$     &$\circ$      &$\circ$      &$\circ$       &$\circ$    &$\circ$    \\\hline
CamVid\cite{camvid}             &2008     &Cambridge        &0.4/-     &701      &$\circ$        &$\circ$     &$\circ$      &$\circ$      &$\circ$       &$\circ$    &$\circ$    \\\hline

\multicolumn{12}{p{2.02\columnwidth}}{\textbf{Notes: } (1) All listed datasets have front facing camera(s). We define Lighting Diversity as whether both daytime and night data were collected, and Weather Diversity if both clear and rainy/snowy/foggy weathers are involved. Driving Behavior studies the driver's facial/posture/steering commands. Designed Choreography refers to designed vehicle-vehicle/pedestrian/bicyclist interactions. (2) In the table, ``$\bullet$" denotes Yes, ``$\circ$" denotes No, and ``-" indicates no information is provided. (3) AA: Ann Arbor, SG: Singapore, CA: California, NY: New York, SF: San Francisco, AZ: Arizona, WA: Washington, AS: ApolloScape, H3D: Honda Research Institute 3D Dataset; HDD: Honda Research Institute Driving Dataset.}
\end{tabular}
\end{table*}

The remainder of this paper is organized as follows: Section \ref{section_related_works} describes related literature. Section \ref{section_vehicle_platform} introduces our vehicle platform and sensors setup. Section \ref{section_collection_labeling} outlines our effort in data calibration, tagging, and labeling. Section \ref{section_analysis_usage} example data analytics. Finally, Section \ref{section_conclusion} points out general conclusions and ongoing/future efforts of our work. 

\section{Related Work} \label{section_related_works}

\subsection{Image Datasets}
Many image datasets have been openly released for AV development. Examples including Imagenet\cite{imagenet} and COCO\cite{coco} provides a seminal starting point for large-scale AI study. CamVid\cite{camvid} offers semantic segmentation for 701 images, and Cityscapes\cite{Cityscapes} captured in 50 cities include pixel-level annotations for 5k images. More recent datasets include Vistas\cite{mapillary}, BDD100k\cite{bdd}, and ApolloScape\cite{appolo}. Some datasets were designed to capture particular diversities/challenges in driving. Vistas and BDD100k target large-scale naturalistic driving from many drivers with wide varieties of weather and lighting, \cite{scnn} focuses on data for lane lines, and \cite{cityperson,nightperson} focuses on pedestrians. 
In the literature there were also efforts that rely exclusively on camera images for AV perception. However, 3D localization using images only is challenging \cite{39,54,46,50}. This leads to a more comprehensive setup of sensors to utilize both semantic (cameras) and ranging sensors (Radar/Lidar/Ibeo). This combination provides better performance or redundancy under hardware failure \cite{nuscenes}. Many datasets released recently include both semantic and ranging sensors. 

\subsection{Multimodal Datasets}
The seminal work that conveys the strength of multimodal sensors is KITTI\cite{kitti}, which provides Lidar scans as well as stereo images and GPS/IMU data. The H3D dataset\cite{h3d} provides annotations in 360$^o$ view, not just the front objects. The KAIST dataset also uses a thermal camera for night time perception \cite{kaist}, Oxford RobotCar studies repeated driving on the same route\cite{oxford}, AppoloScape captures Lidar scans in dense traffic \cite{aslidar}, and nuScenes focuses on 360$^o$ semantic views\cite{nuscenes}. Very recent datasets also includes the work from industrial entities such as Waymo\cite{waymo_dataset} and Lyft\cite{lyft_dataset}. 

\subsection{Driving Behavior Datasets}
The aforementioned datasets primarily focused on data collection for different road environment. We believe another important aspect is the interaction with other road users and driver's steering or speed control inputs. A prominent example of data collection not focusing on AV development is the University of Michigan safety pilot project data\footnote{\url{http://safetypilot.umtri.umich.edu/index.php?content=video}}. This dataset captures vehicle speed, location, and front perception using a MobilEye camera. Many data analysis results have been published\cite{dingzhao_driving_behavior, xianan_paper, xianan_paper_2}. Multimodel datasets also usually include accurate GPS positions, thus providing the possibility of extracting vehicle speed, acceleration, and heading angle for human driver modeling \cite{kitti,nuscenes,oxford}. Datasets focused on human driver behavior also include \cite{honda_dataset, brain4car}.

\subsection{Annotations}
In the literature, different annotation and labeling strategies have been used. For images, 2D bounding boxes\cite{imagenet,bdd}, 3D bounding boxes\cite{kitti,nuscenes,Cityscapes}, and pixel-level segmentation\cite{coco,cam_radar_fusion,bdd} are the most common formats. When (360$^o$) Lidar points are available, 3D bounding boxes may be provided\cite{kitti,aslidar}. For Ibeo and Radar, annotations are usually not provided because the sensor outputs are too sparse or too complicated to annotate.

\begin{figure}[!b]
  \centering
    \includegraphics[width=0.486\textwidth]{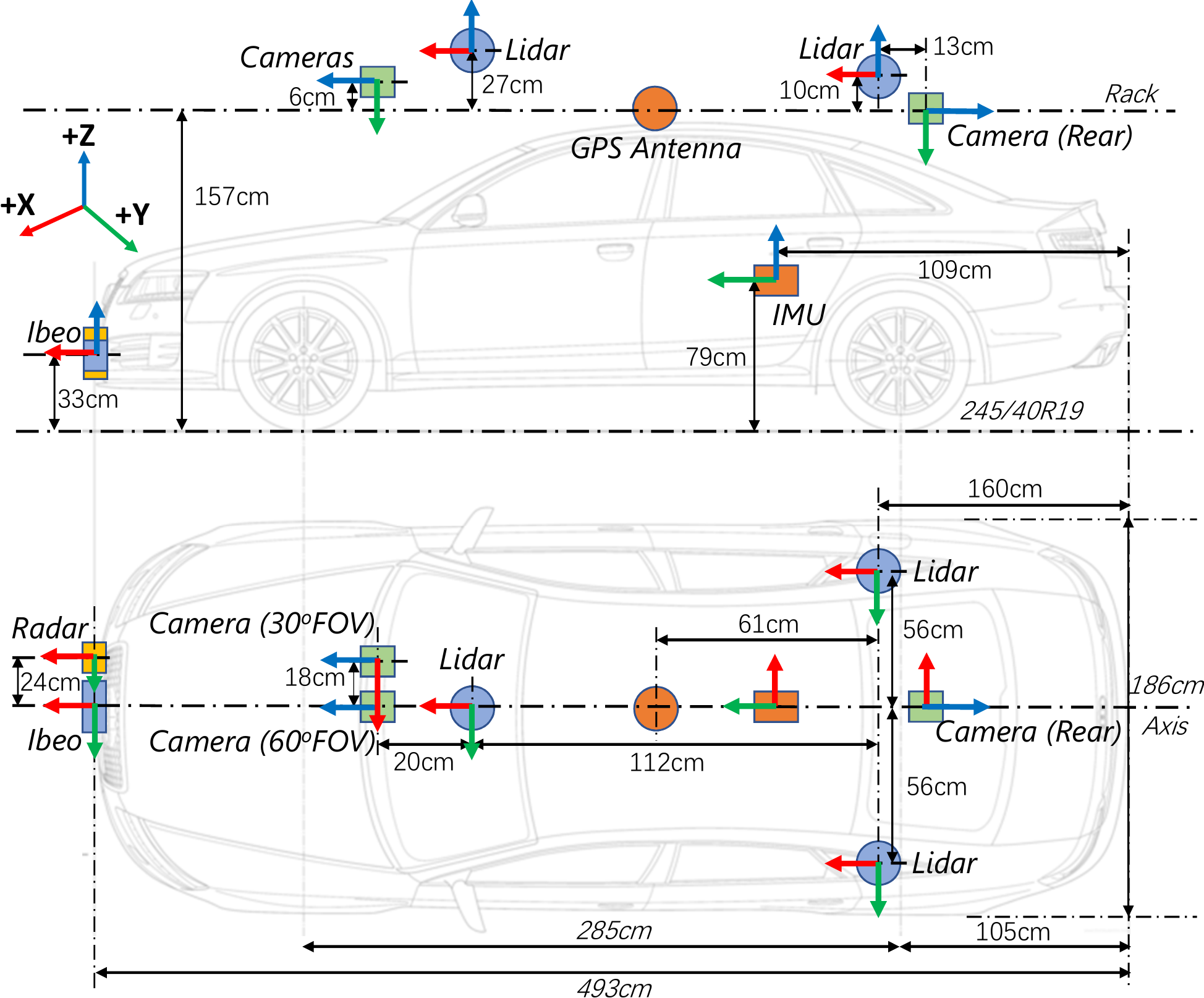}
    \caption{Sensors on the vehicle platform.}
    \label{sensor_setup_plot}
\end{figure}

\begin{figure*}[!t]
  \centering
    \includegraphics[width=2.05\columnwidth]{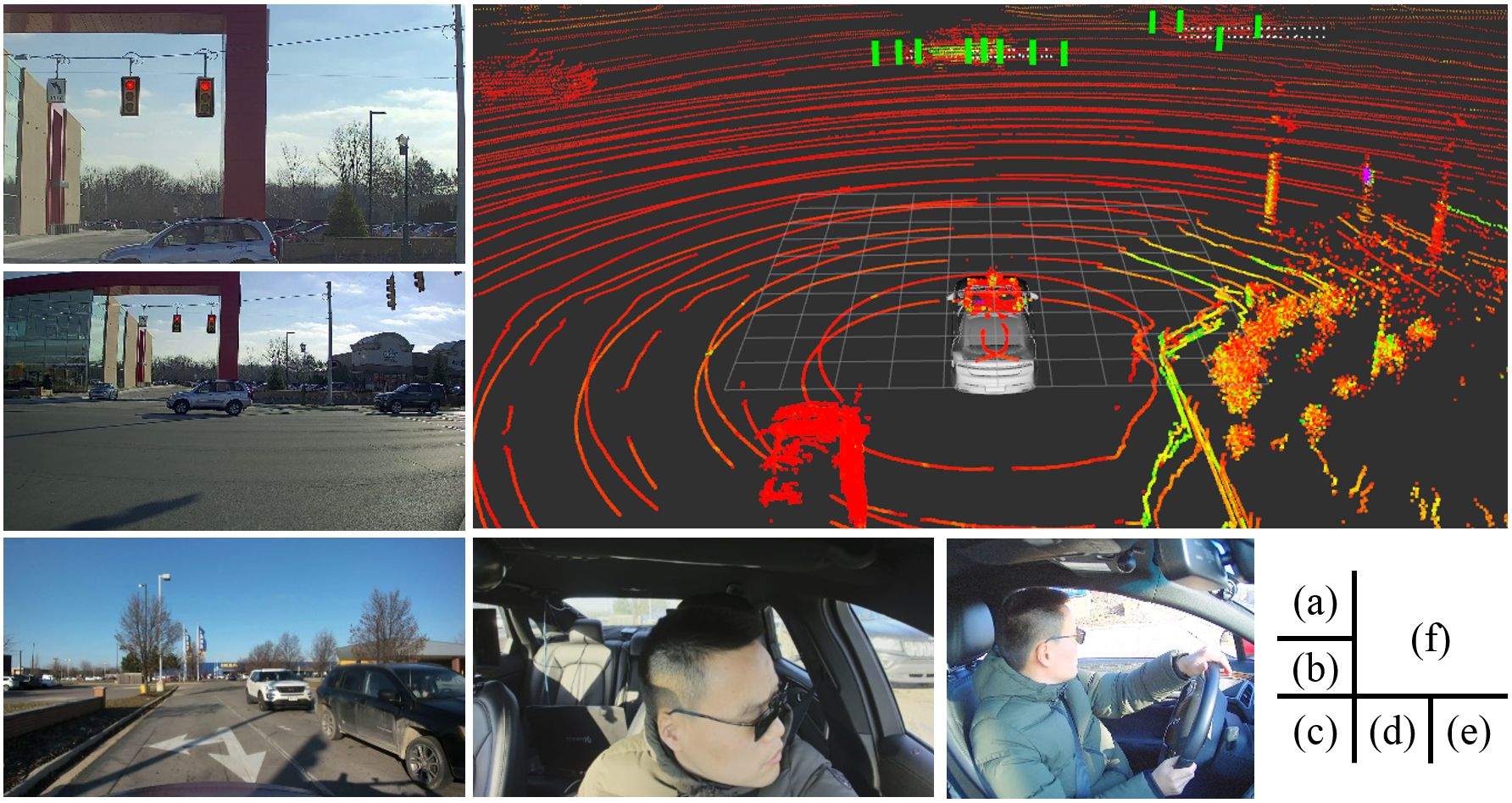}
    \caption{Example sensors outputs: (a) 30$^o$FOV, (b) 60$^o$FOV, (c) Rear, (d) Head/Eyeball, (e) Body pose, (f) Lidar (red/yellow/green point clouds), Radar (green thin cuboids), and Ibeo (white dots)}
    \label{all_sensors_output}
\end{figure*}

\section{Vehicle Platform and Sensors} \label{section_vehicle_platform}
We collect the data manually driving a instrumented Lincoln MKZ. This vehicle is equipped with the following sensors:

\begin{itemize}[leftmargin = 9pt]
\item 3 $\times$ Velodyne Ultra Puck VLP-32C Lidar, horizontal angular resolution 0.2$^o$, vertical 0.33$^o$, range 200m, 10Hz.
\item 2 $\times$ forward-facing cameras, 60$^o$ and 30$^o$FOV, 1080P, 30Hz. 
\item 1 $\times$ backward-facing camera, 90$^o$FOV, 1080P, 30Hz. 
\item 1 $\times$ Cabin pose camera, 1280$\times$1080, 30Hz.
\item 1 $\times$ Cabin head/eyeball camera, 640P, 30Hz.
\item 1 $\times$ Ibeo four beam LUX sensor, horizontal angular resolution 0.25$^o$, vertical 0.8$^o$, range 50m, 25Hz.
\item 1 $\times$ Delphi ESR 2.5 Radar, range 60m, 90$^o$FOV, 20Hz. 
\item 1 $\times$ NovAtel FlexPak6 with IMU-IGM-S1 and 4G cellular for RTK GPS, single antenna, 1Hz. 
\end{itemize}

The locations and example output of the sensors are shown in Fig. \ref{sensor_setup_plot}-\ref{all_sensors_output}. We use two cameras for forward perception, one with wide FOV for general object detection/tracking, the other with narrower FOV for traffic signs and signals. We use a Logitech BRIO camera for backward monitoring which uses a wide FOV. The internal cabin cameras capture the body pose anf head/eyes movement of the human driver. We use three mechanical scanning Lidars, all on the rooftop to capture objects in front, rear left and rear right of the vehicle.

\begin{figure}[t]
  \centering
    \includegraphics[width=0.486\textwidth]{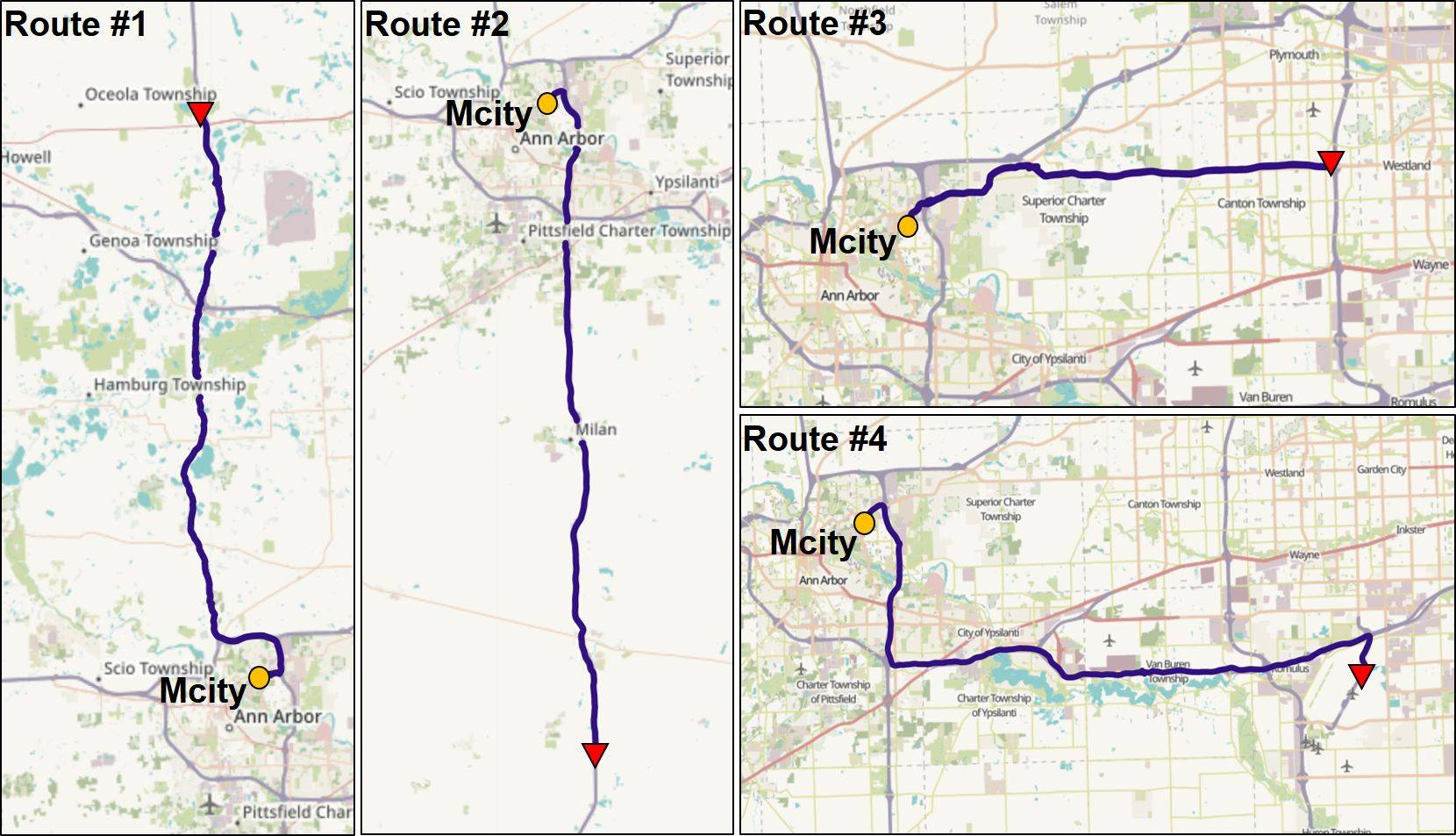}
    \caption{Data collection routes. We start from Mcity, then proceed along US-23 north (\#1), US-23 south (\#2), M-153 east (\#3) and US-94 east (\#4)}
    \label{4_routes_plot}
\end{figure}

\begin{figure*}[!t]
\centering
\begin{subfigure}{.33\linewidth}
    \centering
    \includegraphics[width = \linewidth]{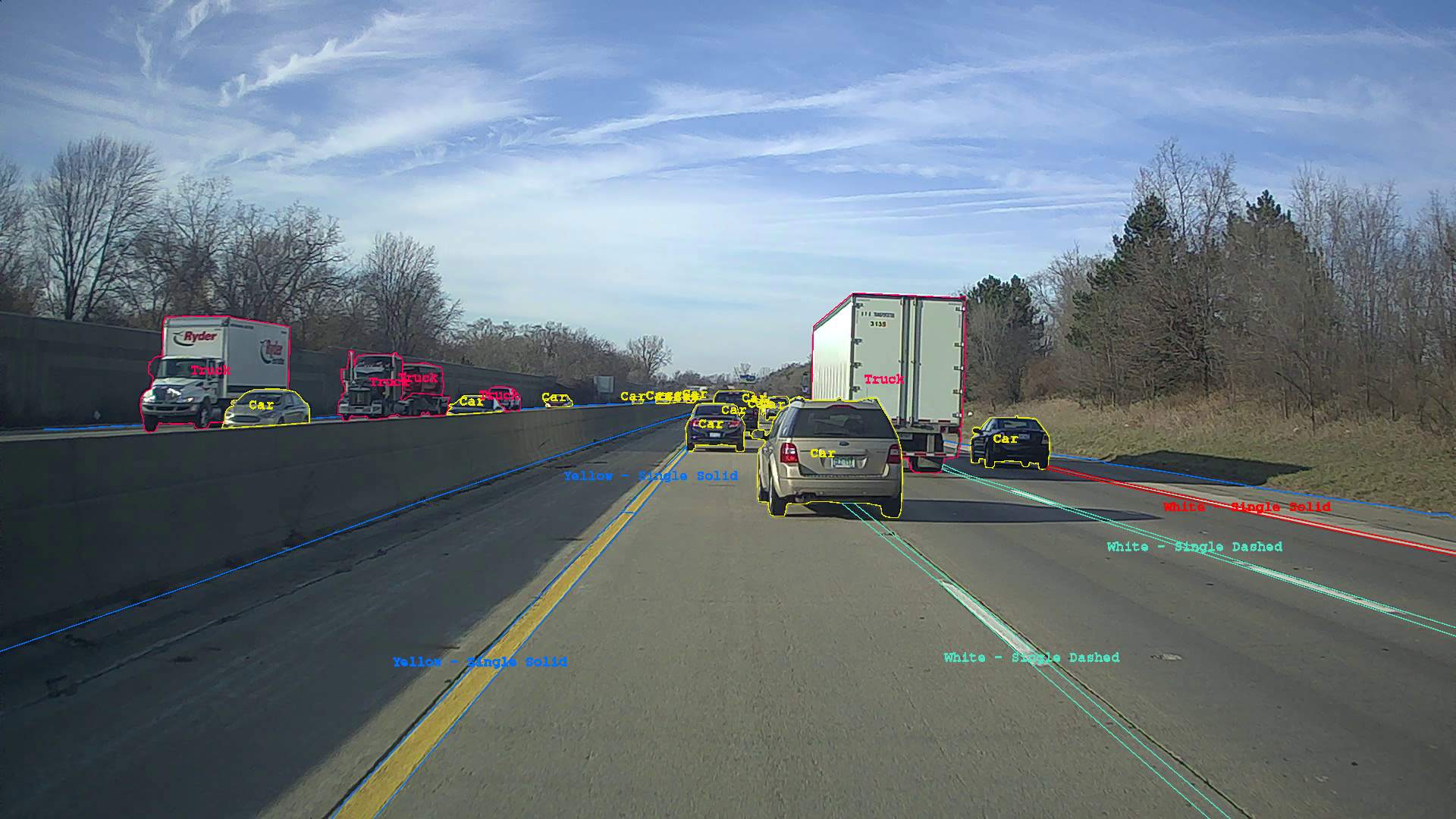}
  \end{subfigure}%
  \hfill
  \begin{subfigure}{.33\linewidth}
    \centering
    \includegraphics[width = \linewidth]{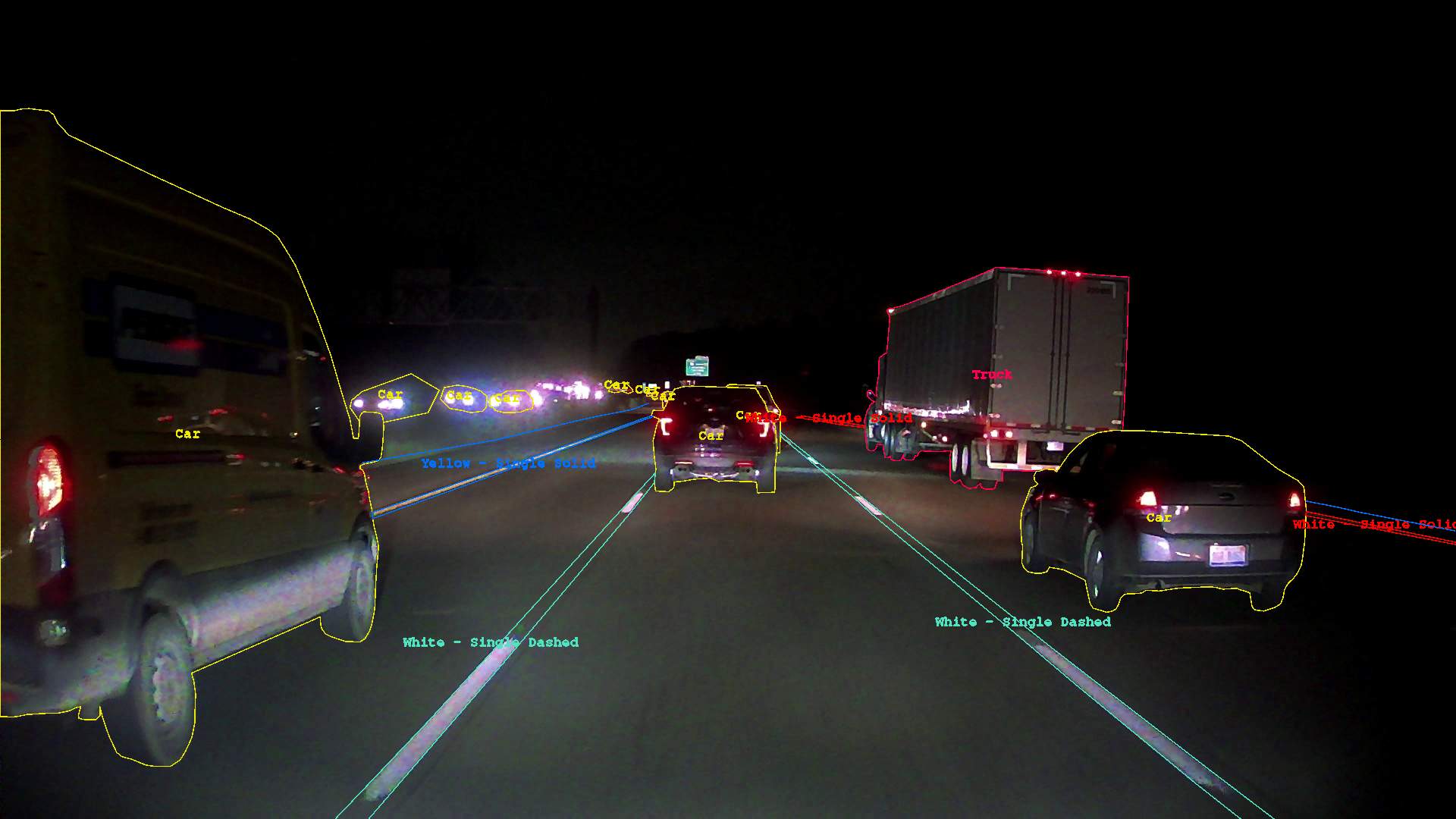}
  \end{subfigure}%
  \hfill
  \begin{subfigure}{.33\linewidth}
    \centering
    \includegraphics[width = \linewidth]{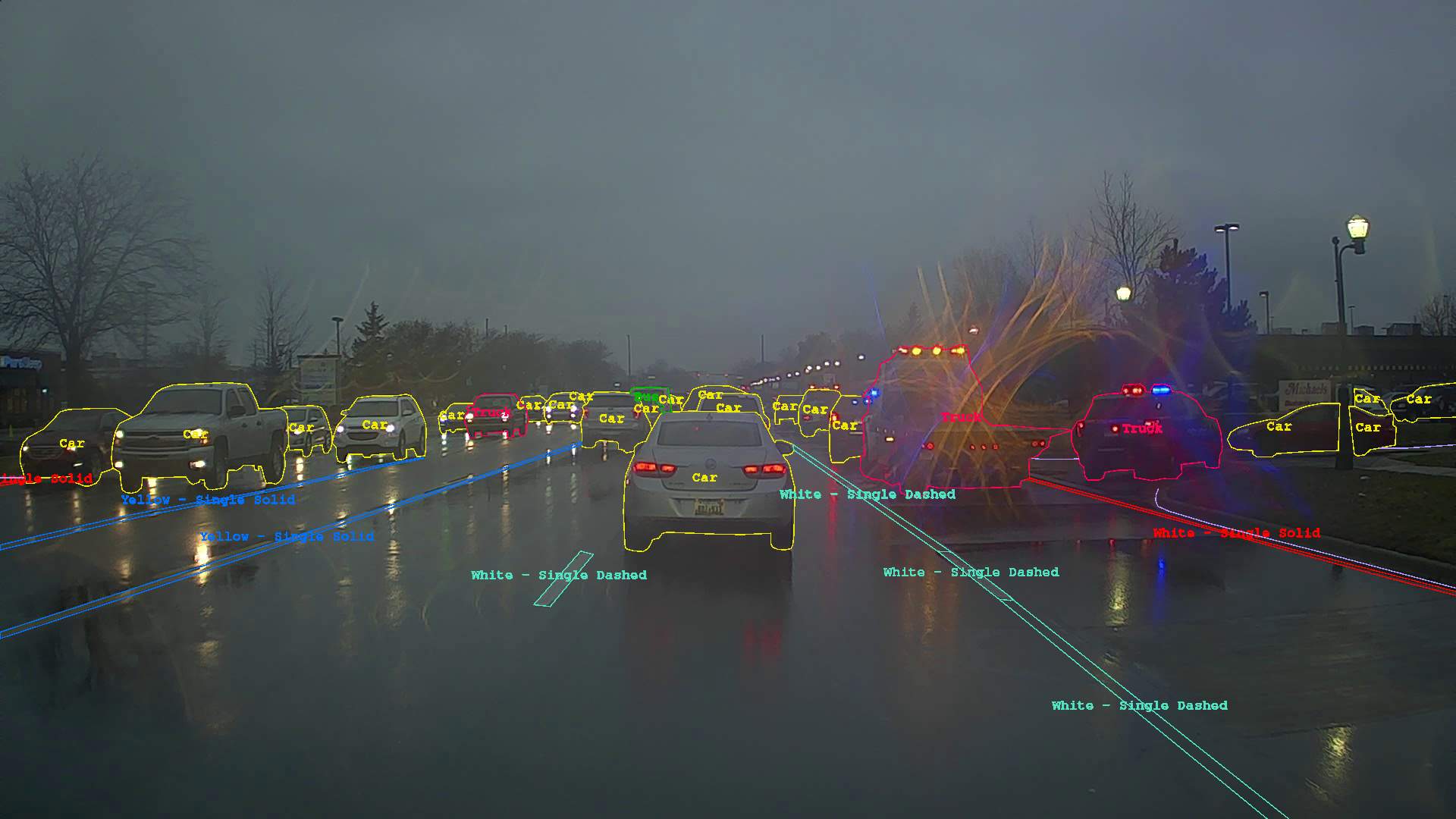}
  \end{subfigure}%
\\ \vspace{3pt}
  \begin{subfigure}{.33\linewidth}
    \centering
    \includegraphics[width = \linewidth]{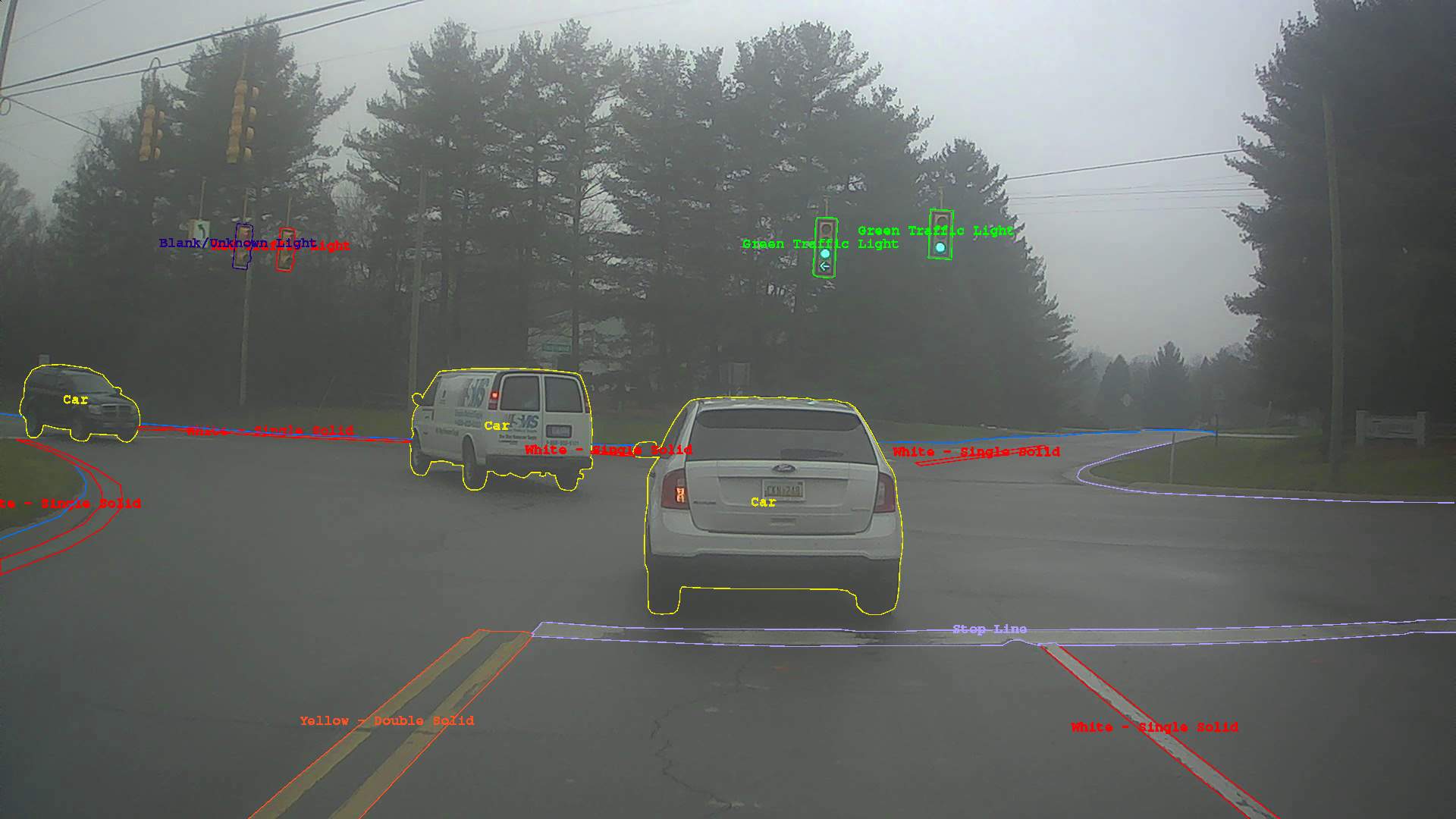}
  \end{subfigure}%
  \hfill
  \begin{subfigure}{.33\linewidth}
    \centering
    \includegraphics[width = \linewidth]{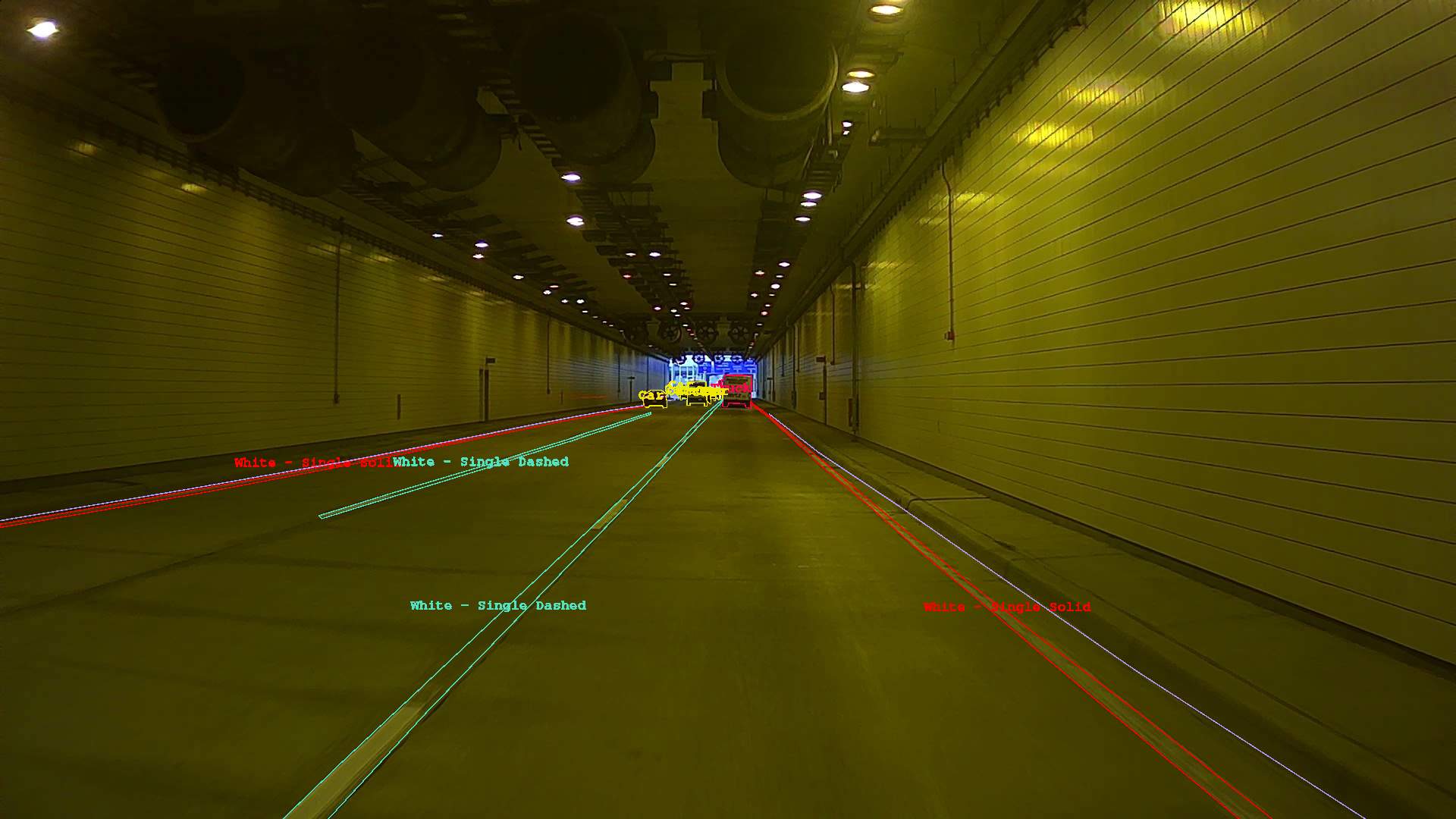}
  \end{subfigure}%
  \hfill
  \begin{subfigure}{.33\linewidth}
    \centering
    \includegraphics[width = \linewidth]{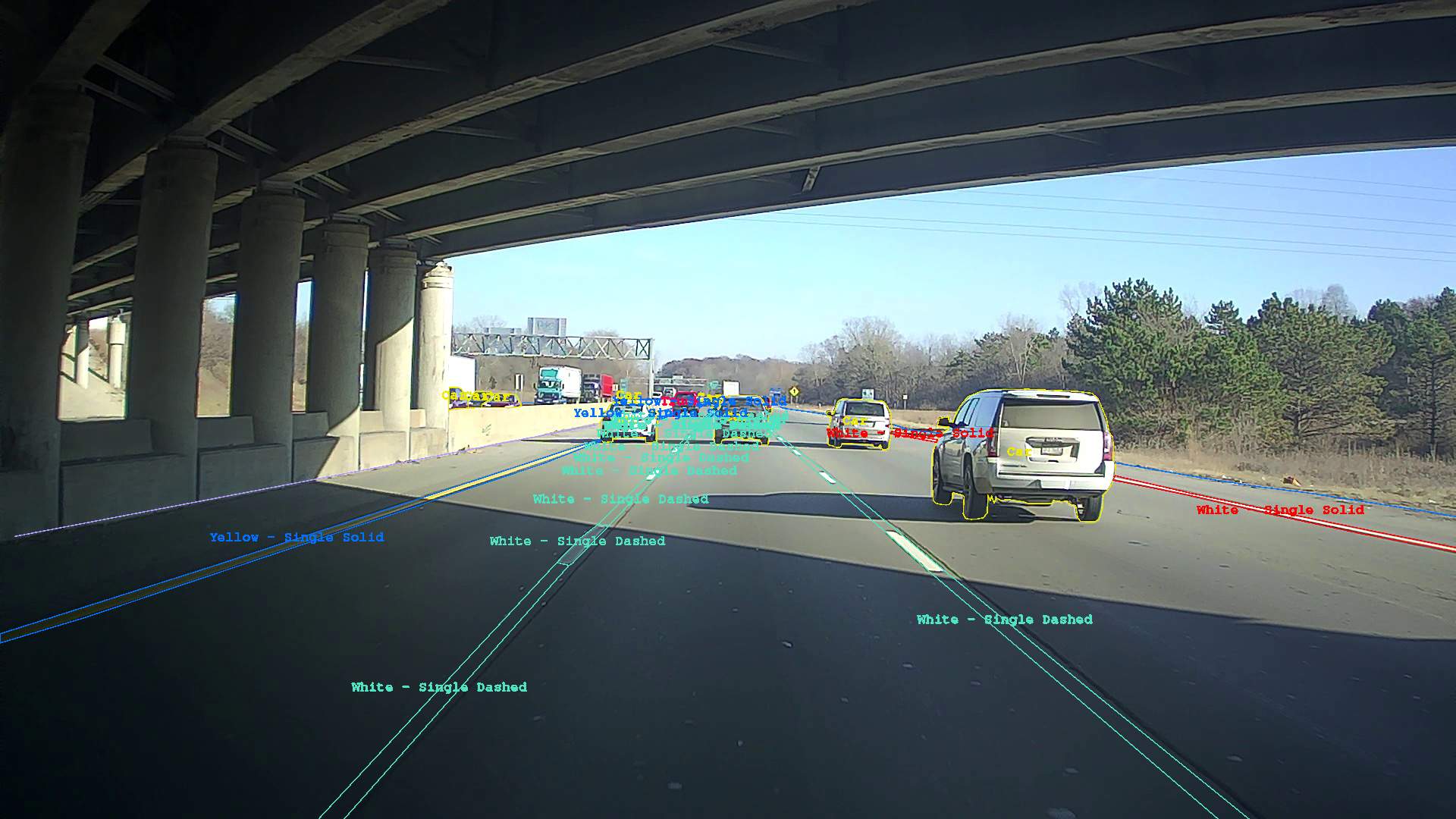}
  \end{subfigure}%
\\ \vspace{3pt}
  \begin{subfigure}{.33\linewidth}
    \centering
    \includegraphics[width = \linewidth]{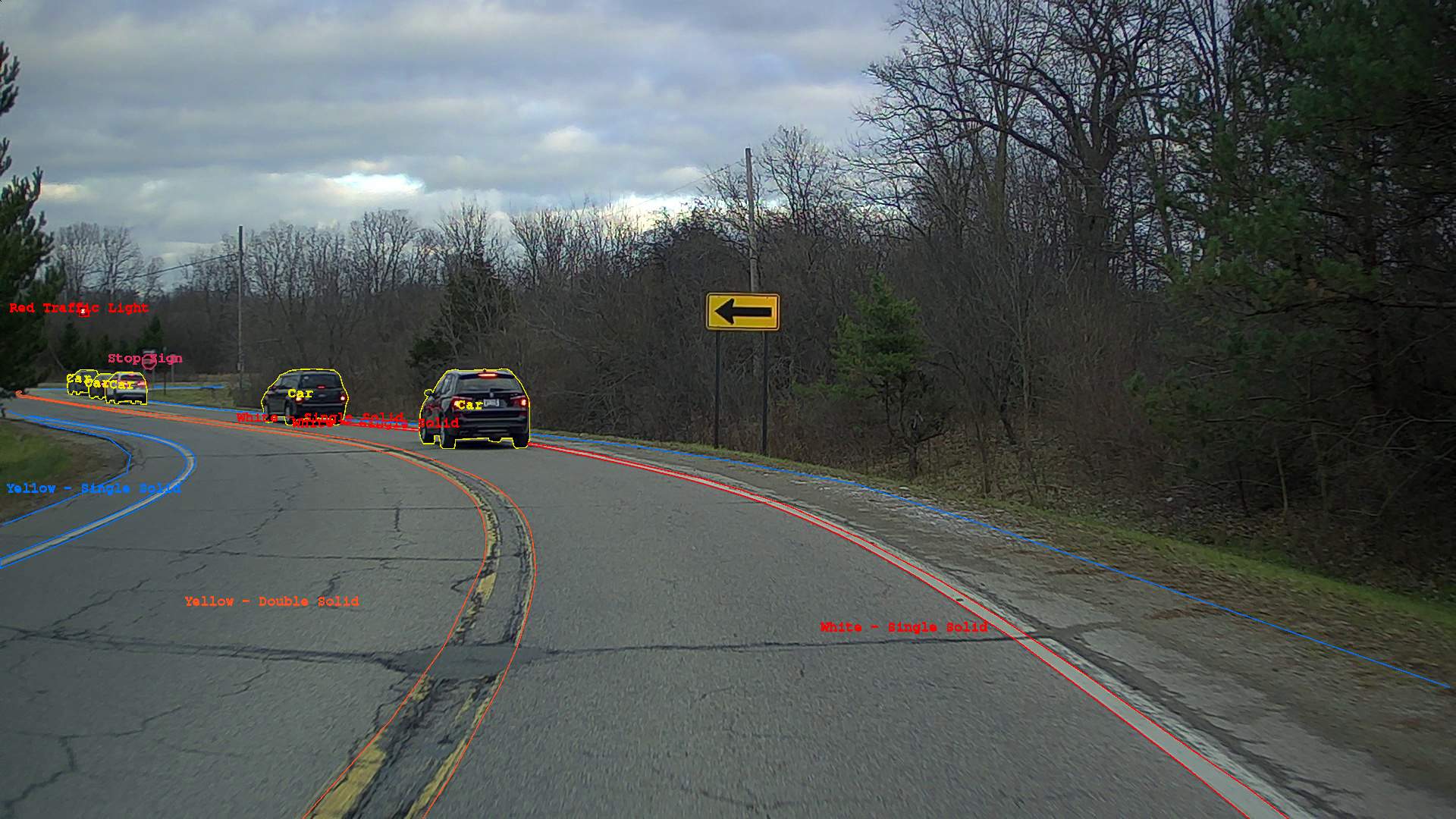}
  \end{subfigure}%
  \hfill
  \begin{subfigure}{.33\linewidth}
    \centering
    \includegraphics[width = \linewidth]{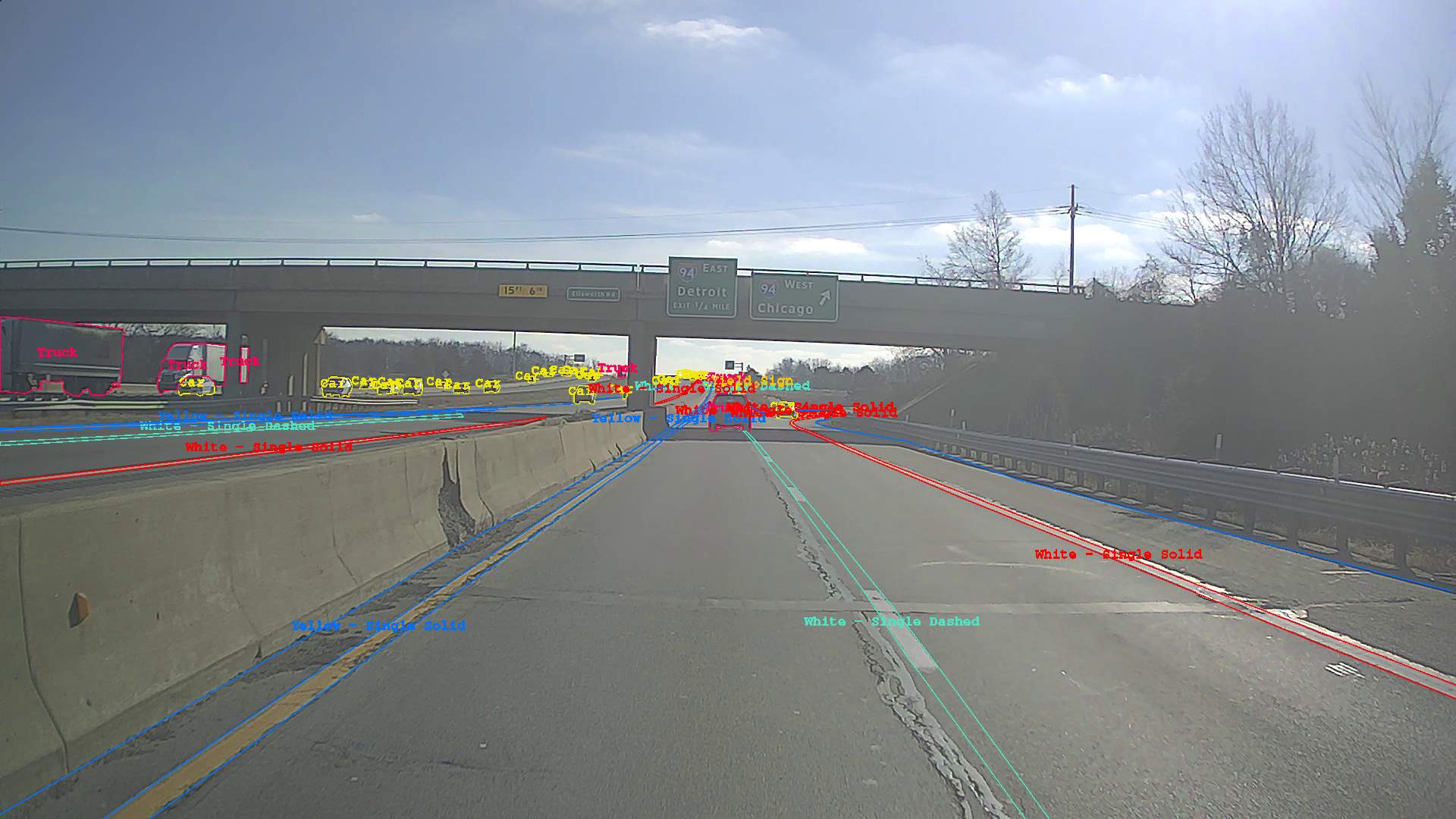}
  \end{subfigure}%
  \hfill
  \begin{subfigure}{.33\linewidth}
    \centering
    \includegraphics[width = \linewidth]{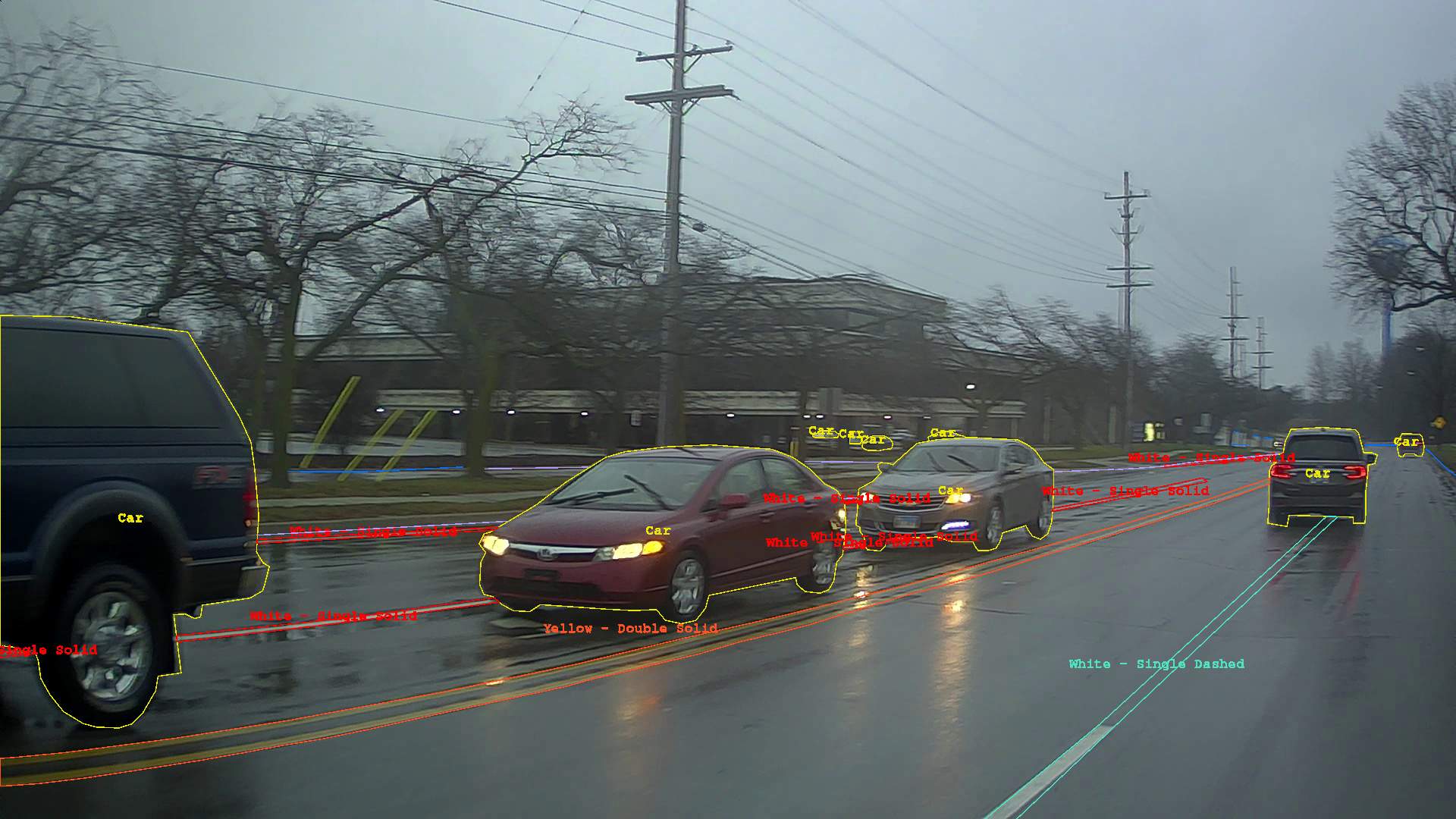}
  \end{subfigure}%
\caption{Data collection and image labeling examples (from the 60$^o$FOV camera). Left to right, top to bottom: clear, night-time, rain, fog, tunnel, bridge, curved road, ramp, intersection.}
\label{collection_labeling_examples_figure}
\end{figure*}

We record all sensors and critical vehicle CAN bus data. The CAN bus reports throttle, brake, and steering commands from the human driver, turn signals, high/low beam state. All the external cameras are connected to a laptop with a GPU, and the videos are recorded via the FFmpeg software. Other sensors (including the head/eyeball movement camera) and the CAN bus data are logged in the ROS formats.  

\section{Data Collection and Annotation\label{section_collection_labeling}}
\subsection{Data Collection Overview}
The data is collected both on open roads and inside Mcity. On open roads, we focus on highways and major local roads. Three human drivers drive manually on these routes with different lighting, weather, road, and traffic conditions. See Fig. \ref{4_routes_plot}-\ref{collection_labeling_examples_figure} for the four routes and an example of the collected scenes. We select routes that take roughly 1 hour round-trip. In total, over 3,000 miles have been covered. In the near future, we plan to focus on urban environments.

\begin{figure*}[!t]
\centering
\begin{subfigure}{2\columnwidth}
\centering
\includegraphics[width=1\textwidth]{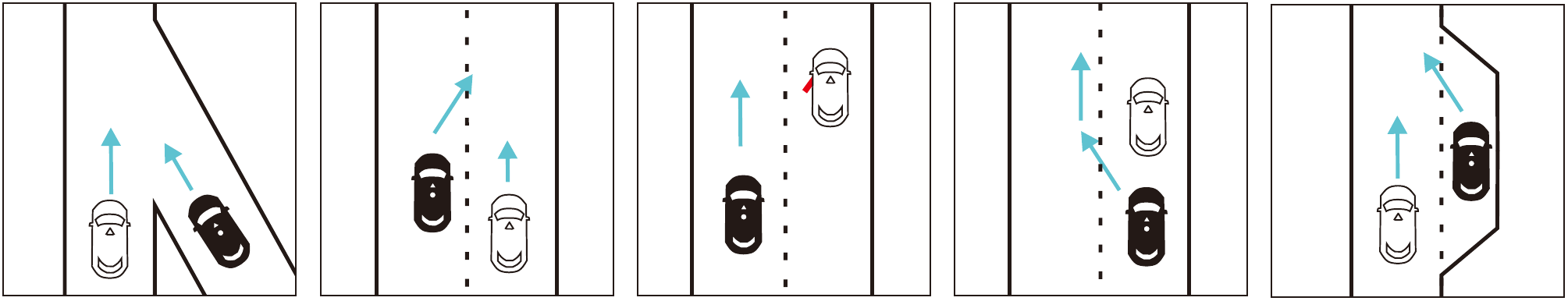}%
\end{subfigure}
\\
\vspace{2pt}
\hspace{0pt}  (1)
\hspace{82pt} (2)
\hspace{83pt} (3)
\hspace{82pt} (4)
\hspace{85pt} (5)
\\
\vspace{2pt}
\begin{subfigure}{1.6\columnwidth}
\centering
\includegraphics[width=1\textwidth]{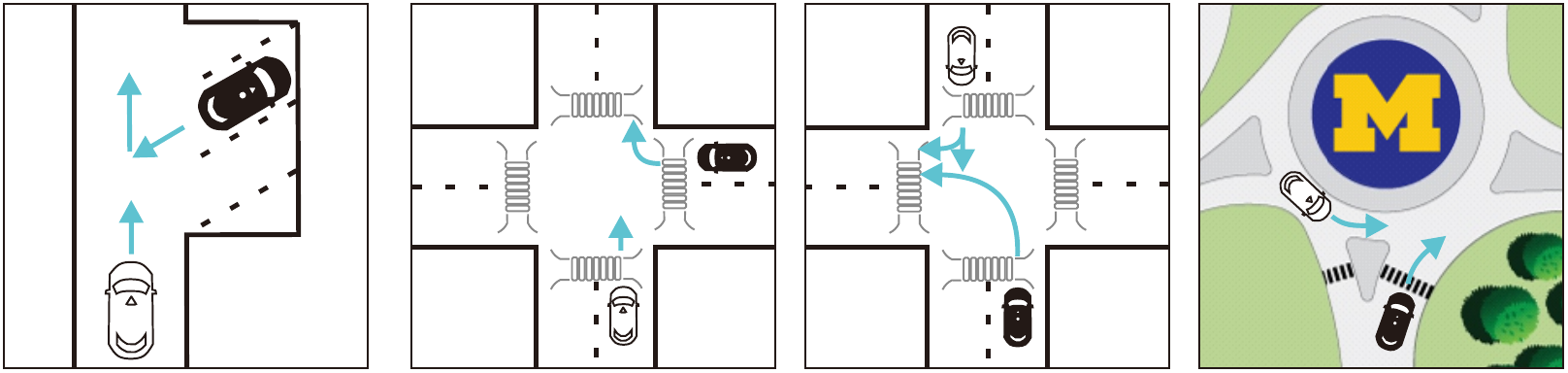}%
\end{subfigure}
\\
\vspace{2pt}
\hspace{0pt}  (6)
\hspace{82pt} (7)
\hspace{83pt} (8)
\hspace{85pt} (9)
\caption{Vehicle-vehicle interactions. (1) Low speed merge (2) Cuts in (3) Door ajar (4) Pass parallel parked vehicle (5) Roadside parked vehicle (6) Angle parked vehicle (7) Right turn (8) Left turn (9) Round-about}
\label{v_v_interactions}
\vspace{2pt}
\begin{subfigure}{2\columnwidth}
\centering
\includegraphics[width=1\textwidth]{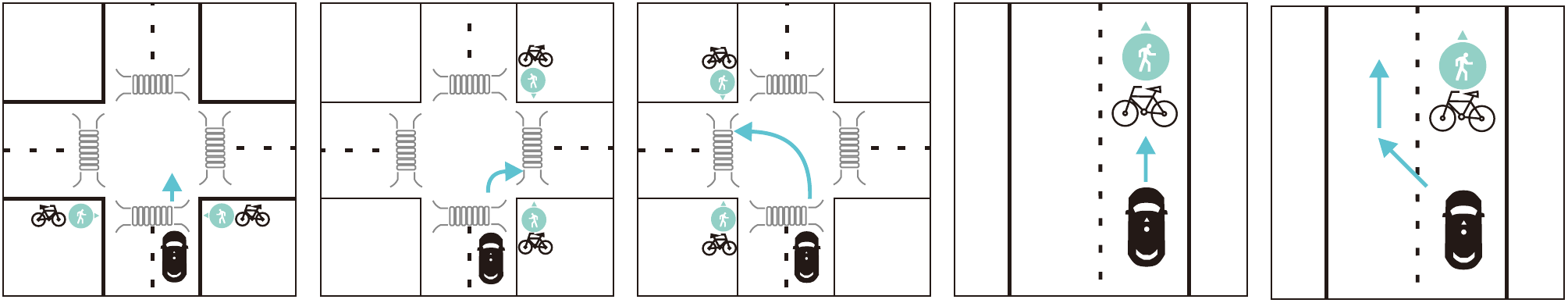}%
\end{subfigure}
\\
\vspace{2pt}
\hspace{0pt}  (1)
\hspace{82pt} (2)
\hspace{83pt} (3)
\hspace{82pt} (4)
\hspace{85pt} (5)
\\
\vspace{2pt}
\begin{subfigure}{1.6\columnwidth}
\centering
\includegraphics[width=1\textwidth]{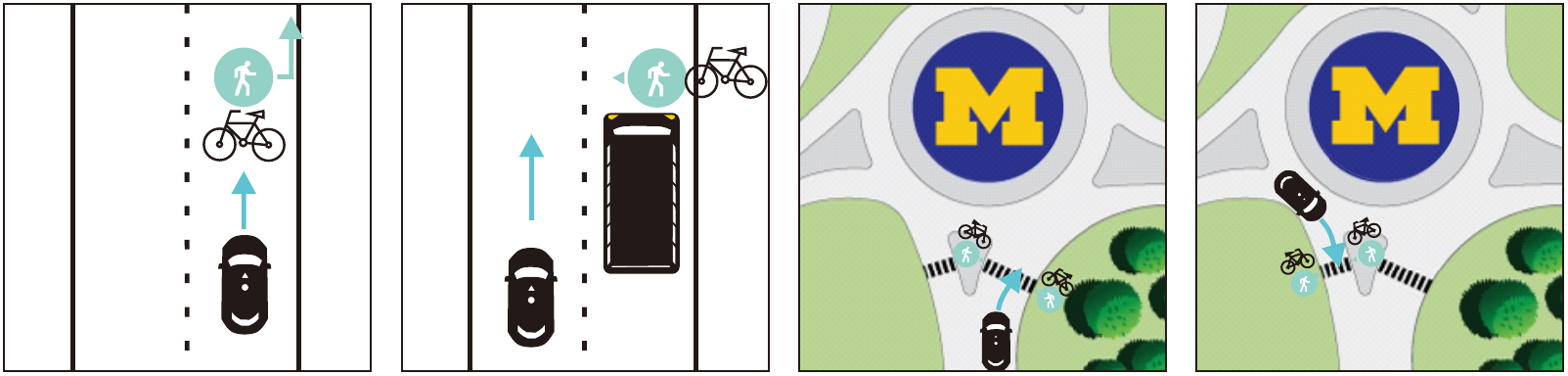}%
\end{subfigure}
\\
\vspace{2pt}
\hspace{0pt}  (6)
\hspace{82pt} (7)
\hspace{83pt} (8)
\hspace{85pt} (9)
\caption{Vehicle-pedestrian/bicyclist interactions. (1) Driving straight at an intersection (2) Right turn at an intersection (3) Left turn at an intersection (4) Follows pedestrian/bicyclist (5) Passing pedestrian/bicyclist on road (6) Pedestrian/Bicyclist yields to vehicle (7) Pedestrian/Bicyclist emerges from behind occlusion (8) Entering round-about (9) Exiting round-about.}
\label{v_p_interactions}
\end{figure*}

The second set of data focuses on designed choreography inside Mcity. We refer particularly to the challenging scenarios in the Mcity ABC testing \cite{mcity_abc}, and design 18 scenarios to study vehicle to vehicle and vehicle to pedestrian/bicyclist interactions, see Fig. \ref{v_v_interactions}--\ref{v_p_interactions} and Table \ref{designed_collection_table}. We record the sensor data wherein both normal (obeying traffic rules) and abnormal (disobeying traffic rules) driving behaviors are involved. We swap the roles of the interacting vehicles when appropriate. We also repeat the collection runs three times for each scenario. 

\begin{table}[!b]
    \centering
    \caption{Designed choreography data collection inside Mcity.}
    \label{designed_collection_table}
    \begin{tabular}{ll}
    \hline
    \hline 
    \multicolumn{2}{c}{\textbf{Vehicle--vehicle interactions}} \\
    \hline
    \textbf{Scenario 1}    &Low speed merge     \\
    \textbf{Scenario 2}    &Vehicle cuts in     \\
    \textbf{Scenario 3}    &Parked vehicle door ajar$^a$     \\
    \textbf{Scenario 4}    &Pass parallel parked vehicle     \\
    \textbf{Scenario 5}    &Roadside parked vehicle start up     \\
    \textbf{Scenario 6}    &Inclined parked vehicle start up     \\
    \textbf{Scenario 7}    &Intersection right turn, other straight     \\
    \textbf{Scenario 8}    &Intersection left turn, other right turn/straight     \\
    \textbf{Scenario 9}    &Vehicle entering round-about     \\
    \hline
    \hline
    \multicolumn{2}{c}{\textbf{Vehicle--pedestrian (P)/bicyclist (B) interactions$^b$}} \\
    \hline 
    \textbf{Scenario 1}         &Vehicle driving straight at intersection     \\
    \textbf{Scenario 2}         &Vehicle right turn at intersection     \\
    \textbf{Scenario 3}         &Vehicle left turn at intersection     \\
    \textbf{Scenario 4\&5}      &Vehicle follows\&passes P/B on road     \\
    \textbf{Scenario 6}         &Pedestrian yields to vehicle driving on road     \\
    \textbf{Scenario 7}         &P/B emerges from behind occlusion     \\
    \textbf{Scenario 8\&9}      &Vehicle entering\&exiting round-about     \\

    \hline
    \multicolumn{2}{p{.9\columnwidth}}{\textbf{$a$: } Other vehicle door ajar, no role swap for the recording MKZ.}\\
    \multicolumn{2}{p{.9\columnwidth}}{\textbf{$b$: } For 1--3, 8, 9, P/B uses the crosswalk to cross road.}
    \end{tabular}
\end{table}

Overall we have collected more than 50 hours of naturalistic driving data covering more than 3,000 miles, and 255 runs for the designed choreography. In total we have roughly 8 TB of ROS files and 3 TB of FFmpeg video.  

\subsection{Synchronization}
The synchronization of the data mainly consists of temporal and spatial calibration. In the temporal calibration, we synchronize using UTC timestamp. For videos recording, we tweak the FFmpeg software to report the UTC timestamp when each frame is written to the disk. For ROS formatted files, the ROS time is equivalent to the UTC timestamp.  

\begin{figure}[!t]
\centering
    \includegraphics[width = \linewidth]{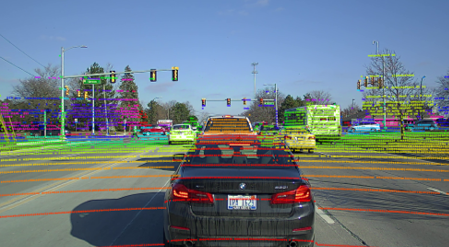}
\caption{Camera-Lidar calibration results.}
\label{cam-lidar-calib}
\end{figure}

\begin{figure}[!t]
\centering
    \includegraphics[width = 1\linewidth]{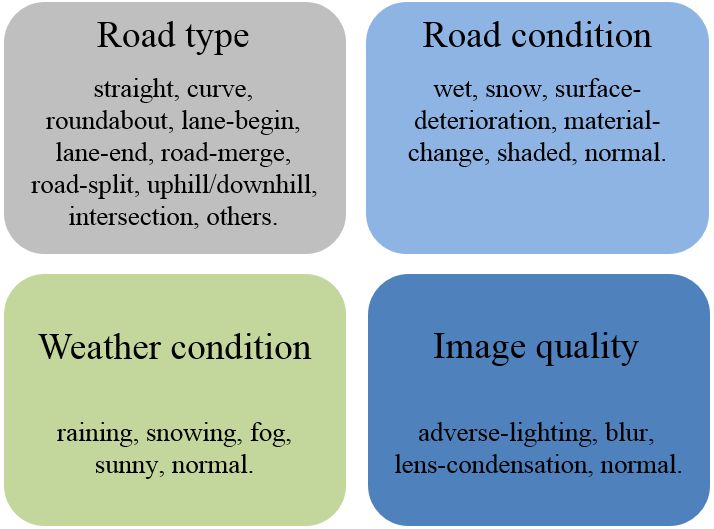}
\caption{Data tagging hierarchy.}
\label{data_tagging_figure}
\end{figure}

The spatial calibration mainly includes camera intrinsic/extrinsic parameters calibration, camera-Lidar, camera-Radar, and camera-Ibeo calibrations. For the camera parameters, we adopt open tools in ROS and use chess boards for the calibration. For camera-Lidar and camera-Ibeo, we follow a methodology similar to KITTI, i.e., using the marker boards wherein manual efforts are needed \cite{yuanxin_calib}. As for camera-Radar calibration, we follow \cite{cam_radar_fusion}. See Fig. \ref{cam-lidar-calib} for an illustration of the camera and front Lidar alignment results. 

\subsection{Data Tagging}
We tag the data (images) for two purposes: for ease of data query, and to balance between the diversity and laborious labeling in the annotation. We devise four tags for each frame: road type, road (surface) condition, weather condition, and image quality. Associated labels are then assigned into each tag. See Fig. \ref{data_tagging_figure} for the tagging hierarchy. More explanation and tag distribution analysis can be found in \ref{section_analysis_usage}.   

\subsection{Data Annotations}
We divide our open road data over the year into 5 stages (batches). Currently we provide results primarily for the front 60$^o$FOV camera. We have annotated more than 17.5k image frames. The annotation class list mainly consists of different objects and traffic signs. Individual files are generated for each frame, illustrating the segmentation boundaries of listed objects/traffic signs. See Fig. \ref{collection_labeling_examples_figure} for example results.  

\begin{figure*}[!h]
\centering
\begin{subfigure}{2.06\columnwidth}
    \centering
    \includegraphics[width = \linewidth]{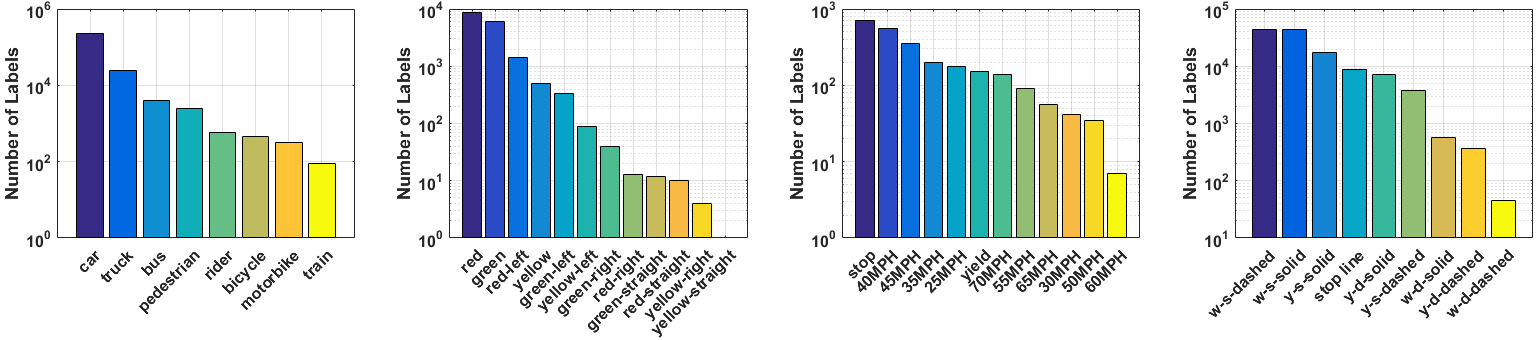}
\end{subfigure}%
\caption{Statistics of different labels in each group. From left to right: object, traffic lights, traffic signs, lanes.}
\label{no_labels_distribution}
\vspace{6pt}
\begin{subfigure}{2.06\columnwidth}
    \centering
    \includegraphics[width = \linewidth]{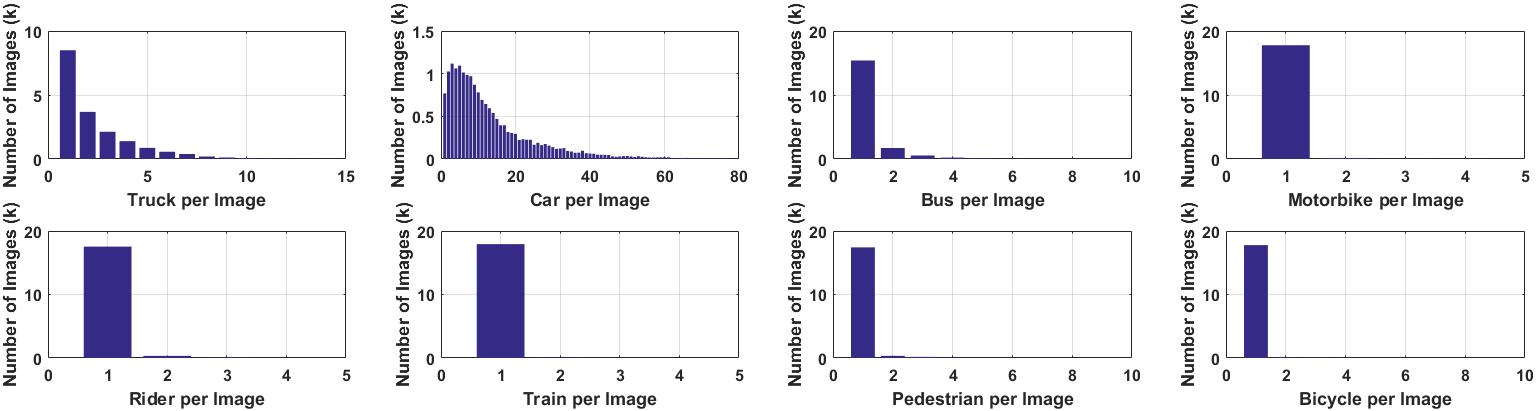}
\end{subfigure}%
\caption{Label density (per image) of the object group.}
\label{no_object_per_image}
\end{figure*}

\begin{figure*}[!h]
\centering
\begin{subfigure}{2.06\columnwidth}
    \centering
    \includegraphics[width = \linewidth]{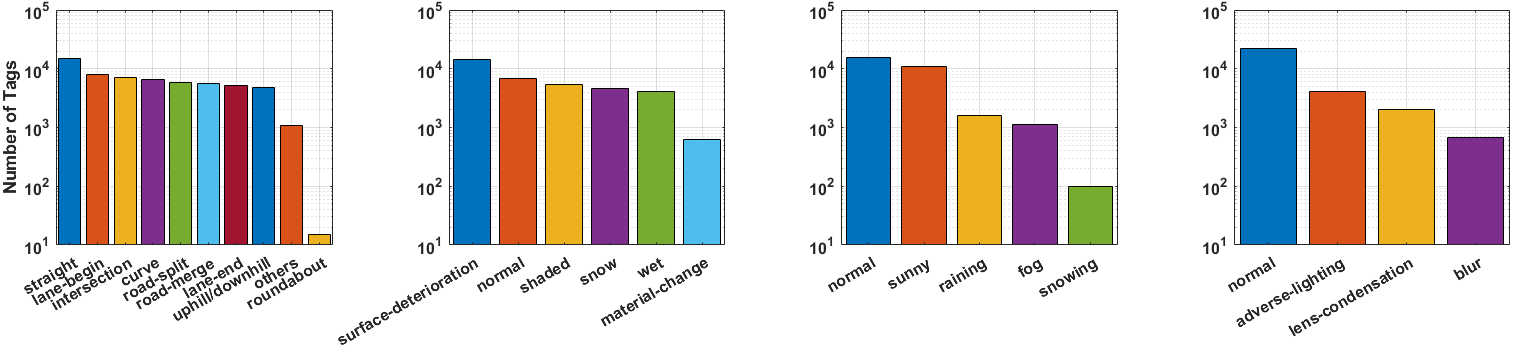}
\end{subfigure}%
\\ \vspace{3pt}
\caption{Data tagging statistics. From left to right: Road type, Road surface, Weather condition, Image quality tags.}
\label{weather_roadsurface_pie}
\end{figure*}

\begin{figure*}[!h]
\centering
    \includegraphics[width = \linewidth]{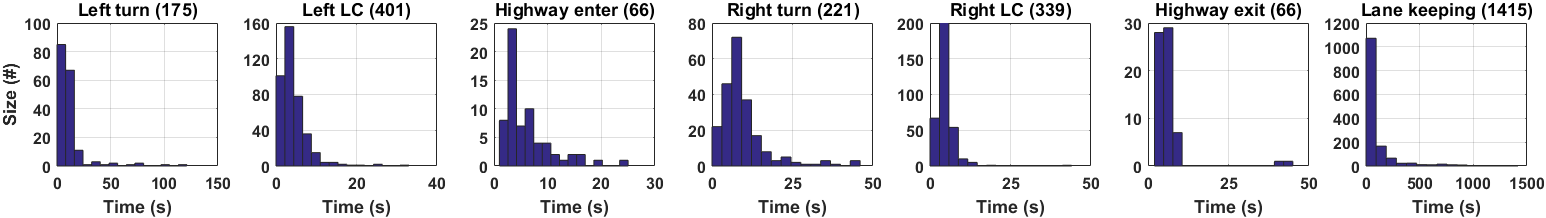}
\caption{Time distribution of the 7 scenarios.}
\label{7_case_distribution_plot}
\end{figure*}

\begin{figure*}[!h]
\centering
    \includegraphics[width = \linewidth]{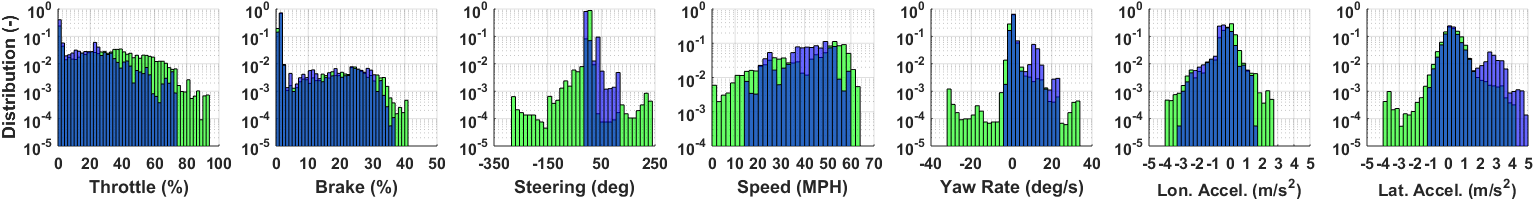}
\caption{Human driving commands and ego motion distribution for highway exit (blue) and right lane change (green) scenarios.}
\label{driving_behave_analysis}
\end{figure*}

\section{Data Analysis and Use} \label{section_analysis_usage}
\subsection{Image Annotations}
We perform diversity analysis for our current image. We divide our annotations into 4 groups, i.e. objects, traffic lights, traffic signs, and lanes. Fig. \ref{no_labels_distribution} shows the statistics of different labels in each group, and Fig. \ref{no_object_per_image} shows example results of number of labels in the object group. For the other three annotation groups, we label traffic lights and traffic signs, visually discernible lane lines/markers/road curbs/stop line, and labels for each lane line segments, which is among the most elaborated datasets we are aware of.   

Statistics pertaining to the images tagging are shown in Fig. \ref{weather_roadsurface_pie}. For weather conditions, most of the data were captured in normal or sunny weathers. However, rainy, foggy, and snowing days were also included. Road condition mainly depicts the lighting/friction condition on the road surface. We mainly consider surface deterioration, material change, snow coverage on the road in our tagging. Statistics of image quality and road types tagging are shown in Fig. \ref{weather_roadsurface_pie}. While many other datasets include data only when the camera works perfectly, poor image quality due to weather or hardware malfunction should be considered. We include both normal, adverse-lighting, lens-condensation, and blur images. Road type describes the shape of collected road/lane lines. See the statistics in Fig. \ref{weather_roadsurface_pie}. The tagging for such property is also quite elaborated among all open datasets. 

\subsection{Driving Behaviors}
Our data includes open naturalistic driving and designed choreography inside the Mcity test facility. For latter the behaviors are illustrated in Table \ref{designed_collection_table} and Fig. \ref{v_v_interactions}-\ref{v_p_interactions}, the analysis in this section focuses on the open road data. Following \cite{kitti, honda_dataset}, we discuss ego motions of the recording vehicle. However, we analyze the driving behaviors separately for different scenarios. Following the previous research efforts for highway entrance/exit in \cite{highway_enter_1, highway_exit}, lane change in \cite{LC_2}, and intersection interactions in \cite{Intersection_1}, we split the recorded data in each run into 7 different scenarios, i.e. left turn, left lane change (LC), ramp entrance, right turn, right LC, highway exit, and lane keeping. We then organize the data following these scenarios. See the distribution plot of the 7 scenarios in Fig. \ref{7_case_distribution_plot}. We also mark the total size (number) of each scenario we have recorded in the figure. To our best knowledge, our dataset is the only one that organizes according to driving scenarios. 

The results of human commands and vehicle motion for right LC and highway exit scenarios can be seen in Fig. \ref{driving_behave_analysis}. Although both scenarios should use right turning signal, the distributions for the states and commands are distinctively separated. This indicates the need to organize data based on driving scenarios. We are currently annotating the collected data to summarize the perception data.  

\subsection{The Complete Driving Flow}

\begin{figure*}[!h]
\centering
\begin{subfigure}{\linewidth}
\centering
\includegraphics[width=1\textwidth]{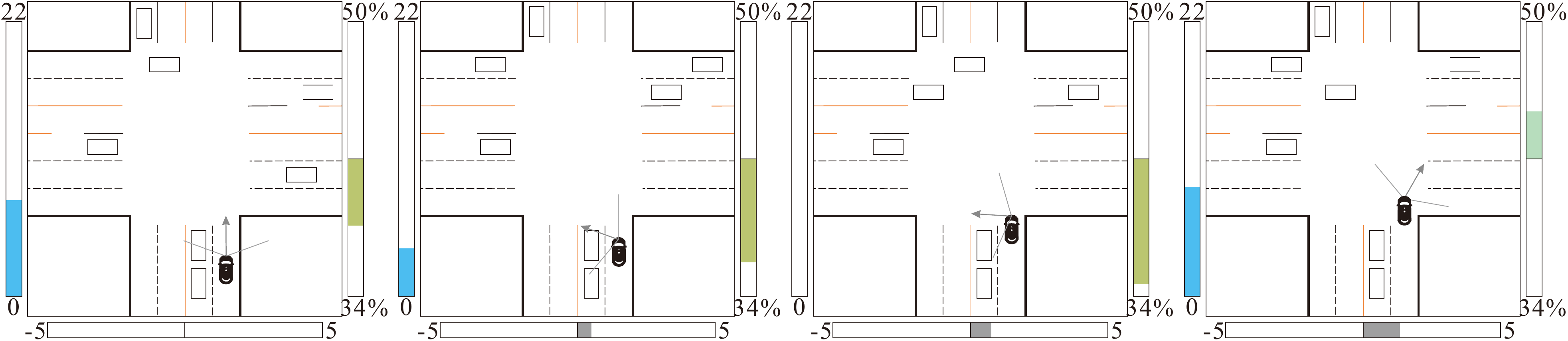}%
\end{subfigure}
\\
\vspace{2pt}
\hspace{-10.5pt}  (1)
\hspace{111pt}   (2)
\hspace{110.5pt} (3)
\hspace{110pt}   (4)
\\
\vspace{2pt}
\begin{subfigure}{\linewidth}
\centering
\includegraphics[width=1\textwidth]{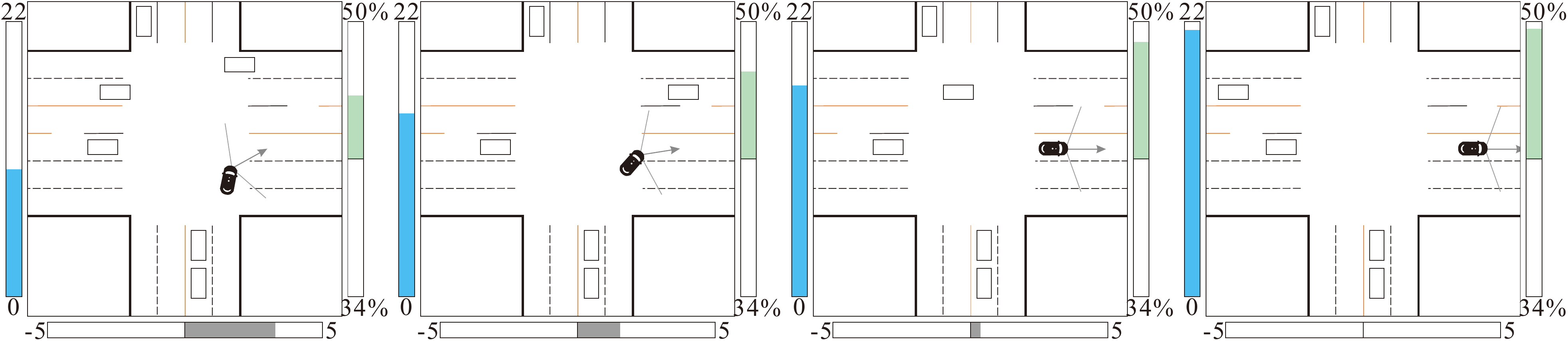}%
\end{subfigure}
\\
\vspace{2pt}
\hspace{-10.5pt}  (5)
\hspace{111pt}   (6)
\hspace{110.5pt} (7)
\hspace{110pt}   (8)
\caption{Obeying traffic rules: a complete trip of unprotected right turn on open roads. This figure depicts the same scene as Fig. \ref{all_sensors_output}. The driver (black vehicle) coasts down to the intersection (1-2), brakes to a full stop at the stop line, and turns head to the left to check incoming traffics (3). Once a safe gap is found, the driver turns head back, steers the vehicle (4-6), and accelerates (7-8). Note we draw the vehicle state and driver's commands in each plot; bar on the left: speed (MPH), right: normalized throttle (up)/brake (down), bottom: steering angle (rad, positive to the right).}
\label{complete_driving_flow_open}
\end{figure*}

\begin{figure*}[!h]
\centering
\begin{subfigure}{\linewidth}
\centering
\includegraphics[width=1\textwidth]{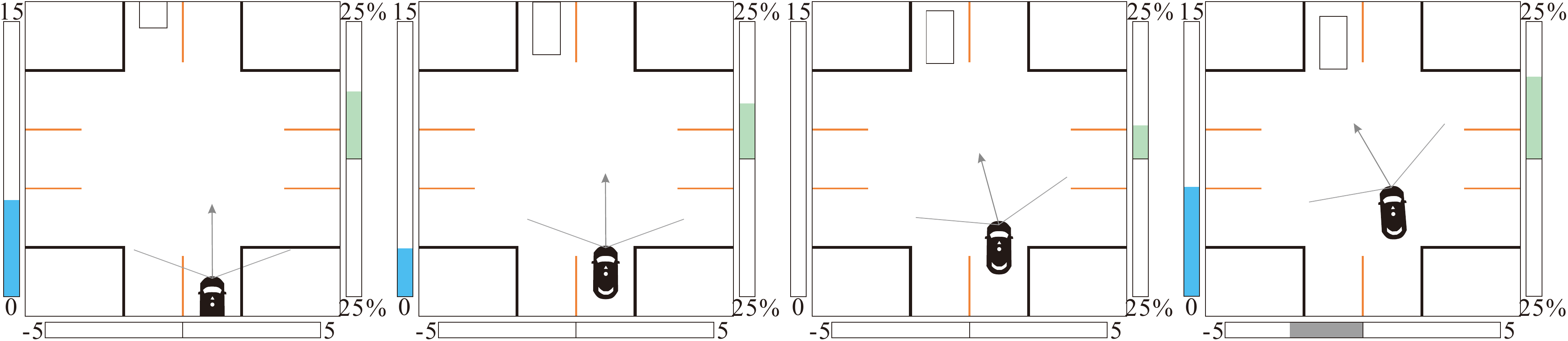}%
\end{subfigure}
\\
\vspace{2pt}
\hspace{-10.5pt}  (1)
\hspace{111pt}   (2)
\hspace{110.5pt} (3)
\hspace{110pt}   (4)
\\
\vspace{2pt}
\begin{subfigure}{\linewidth}
\centering
\includegraphics[width=1\textwidth]{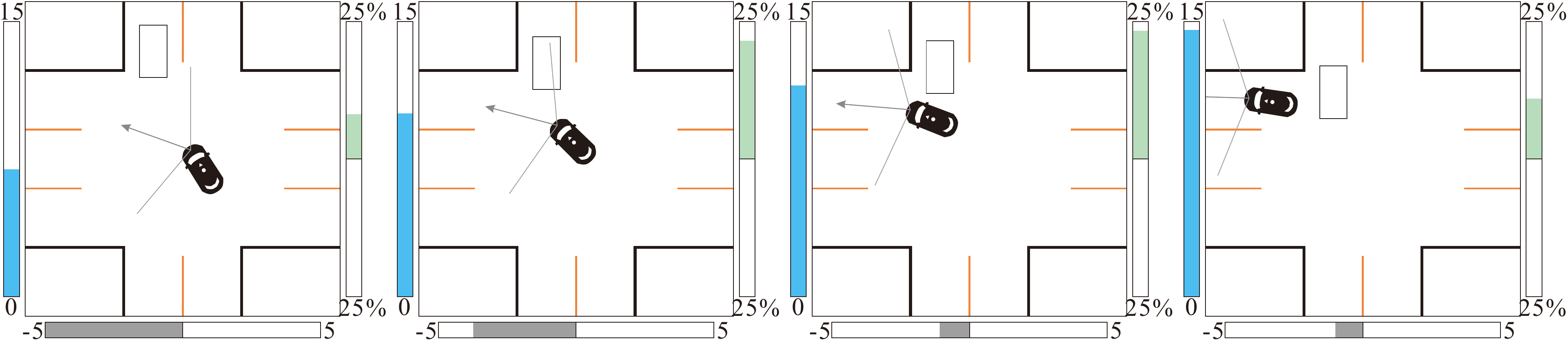}%
\end{subfigure}
\\
\vspace{2pt}
\hspace{-10.5pt}  (5)
\hspace{111pt}   (6)
\hspace{110.5pt} (7)
\hspace{110pt}   (8)
\caption{Disobeying traffic rules: a complete trip of an unprotected left turn inside Mcity. The human driver (black vehicle) did not yield to the oncoming vehicle. In (1-3), the driver coasts down to the intersection, turns head, visually checks the incoming traffics and evaluates the safe gap. In (4-7), the driver does not yield, accelerates and turns. In (8), both vehicles accelerate to leave the intersection. In this turning, the gap was small: the oncoming vehicle had to brake to avoid collision. Driver's commands are: bar on the left: speed (MPH), right: normalized throttle (up)/brake (down), bottom: steering angle (rad, positive right).}
\label{complete_driving_flow_mcity}
\end{figure*}

Our data also provides the complete flow of human driver's actions. We show two examples in Fig. \ref{complete_driving_flow_open}-\ref{complete_driving_flow_mcity}. Fig. \ref{complete_driving_flow_open} depicts an unprotected right turn on open roads. The human driver catches a safe gap to make the turn. In Fig. \ref{complete_driving_flow_mcity}, we illustrate a trip wherein the driver disobeys traffic rules in an unprotected left turn. In both figures, we record and plot the visual perception direction (gray arrow), throttle, brake, and steering commands; the ego vehicle speed is also shown. We believe that in addition to be efficient and safe, being naturalistic is also a desired trait for AV. The complete data capture will be useful for such analysis. 

\section{Conclusion and Future Works}\label{section_conclusion}
This paper presents the ongoing data collection effort at Mcity. Compared to existing datasets, our data is complete with all commonly used sensor types. We collect the data both on open roads naturalistically and inside the Mcity test facility with designed choreography. We perform preliminary analysis on our data, which use tags to indicate different driving scenarios and conditions.   

\section*{Acknowledgement}
We want to thank many students/engineers at Mcity for the vehicle platform development and their thoughtful suggestions on data analysis. We also thank Seres for providing the vehicle and funding the project, and Might AI for image labeling.  
\bibliographystyle{unsrt}
\bibliography{bibliography.bib}
\end{document}